\documentclass{article}
\usepackage{amsmath}
\usepackage{tcolorbox}
\usepackage{multirow}
 \usepackage[preprint]{neurips_2026}

\usepackage[utf8]{inputenc} 
\usepackage[T1]{fontenc}    
\usepackage{hyperref}       
\usepackage{url}            
\usepackage{booktabs}       
\usepackage{amsfonts}       
\usepackage{nicefrac}       
\usepackage{microtype}      
\usepackage{xcolor}         
\usepackage{wrapfig}
\usepackage{verbatim}
\usepackage{float}
\usepackage{enumitem}

\title{AsymVLM: Asymmetric Token Pruning for Efficient Vision-Language Model Inference}

\author{
  Yilin Feng\thanks{Equal contribution} \\
  The Pennsylvania State University\\
  \texttt{ypf5071@psu.edu} \\
  \And
  Ahmed Burak Gulhan\footnotemark[1] \\
  The Pennsylvania State University \\
  \texttt{gulhan@psu.edu} \\
  \AND
  Mahmut Taylan Kandemir \\
  The Pennsylvania State University \\
  \texttt{mtk2@psu.edu} \\
}

\begin{document}

\maketitle

\begin{abstract} 
Vision-Language Models (VLMs) process thousands of visual tokens per image alongside comparatively few text tokens, yet existing compression methods treat both modalities uniformly. We observe that the two modalities have fundamentally different properties: vision tokens are spatially redundant and dominate prefill, while text tokens are causally dependent and accumulate during decoding. Based on this asymmetry, we propose and empirically evaluate {\bf AsymVLM}, which applies aggressive pruning to vision tokens before prefill using a learned importance scorer with per-sample adaptive budgeting, and temporal threshold-based eviction to text tokens only when they exceed a fixed budget. Our experiments indicate that AsymVLM achieves the highest FLOPs savings (up to 54\%) among state-of-the-art methods while outperforming existing approaches by 2--3\% on document and chart understanding tasks where visual information is spatially localized and query-specific, and maintaining competitive accuracy on holistic benchmarks. In text-dominated scenarios, our eviction strategy substantially outperforms standard LLM cache compression methods by adapting to the short-context nature of VLM.
\end{abstract} 

\vspace{-10pt}
\section{Introduction}
\vspace{-5pt}
Vision-language models (VLMs) have achieved remarkable success across a wide range of multimodal tasks, from visual question answering and document understanding to image captioning and embodied reasoning~\citep{liu2023llava, liu2024improved, bai2023qwenvl, abdin2024phi3}. By integrating pretrained vision encoders with large language models (LLMs), architectures such as LLaVA \citep{liu2023llava, liu2024improved}, Phi-3-Vision \citep{abdin2024phi3}, and Qwen-VL \citep{bai2023qwenvl} inherit the powerful reasoning capabilities of their language model backbones while extending them to visual input. However, this integration introduces significant computational and memory challenges: each image is encoded into hundreds or even thousands of visual tokens; as the context length increases, the size of the Key-Value (KV) cache grows linearly, thereby becoming a major bottleneck affecting inference latency and GPU memory consumption.

A growing body of work has explored visual token reduction to address this bottleneck. For example, FastV \citep{chen2024fastv} and SparseVLM \citep{zhang2024sparsevlm} prune vision tokens based on attention patterns within the LLM layers, while FlashVLM \citep{cai2025flashvlm} and VisPruner \citep{zhang2025vispruner} perform one-shot pruning at the encoder--decoder boundary using visual saliency or cross-modal similarity. These methods score vision tokens based on semantic relevance, retaining patches that are visually salient or textually aligned. However, semantic similarity alone does not fully capture a token's contribution to the model's final output. A vision token may appear semantically distant from the query, yet still influence the last hidden states through indirect attention pathways across layers. Furthermore, existing methods apply a fixed, uniform pruning ratio across all samples regardless of how visual information is distributed
in the image, and address only the vision modality while leaving text token management to standard autoregressive caching.

After conducting a comprehensive analysis of VLM architectures and major visual benchmarks, we identify two key observations that motivate a modality-aware token compression approach:

\textbf{First, vision and text tokens differ in structural properties, and the memory bottleneck shifts between them depending on the task.} Vision tokens are spatial patches with no causal dependency among them, exhibiting high spatial redundancy that permits independent removal. Text tokens, on the other hand, are generated autoregressively with strict causal ordering, where evicting any token risks corrupting all subsequent representations.  In typical VLM configurations, vision tokens dominate the input (e.g., 2{,}500 vision tokens vs.\ 20--30 text tokens in Phi-3.5-Vision in various VLM benchmarks), making them the primary memory bottleneck during prefill.  However, in free-form generation and multi-turn dialogue settings, text tokens accumulate rapidly during decoding and can eventually exceed vision tokens in memory consumption, thereby shifting the system bottleneck from the prefill stage to decoding. 

\textbf{Second, the optimal vision token budget varies substantially across samples.} Different inputs demand fundamentally different levels of visual coverage depending on where task-relevant information resides in the image. Consider two document understanding examples: When asked ``What is the underlined heading just above the table?'', the answer is localized to a small region, and most of the image can be safely discarded.  On the other hand, a question like ``How many boxed illustrations are there?'' demands global scanning of the entire image to count scattered elements. A fixed pruning ratio cannot accommodate this spectrum because it either wastes computation in the first case or destroys critical information in the third.

\begin{wrapfigure}{r}{0.6\textwidth}
\vspace{-15pt}
\label{fig:overview}
  \begin{center}
    \includegraphics[width=0.6\textwidth]{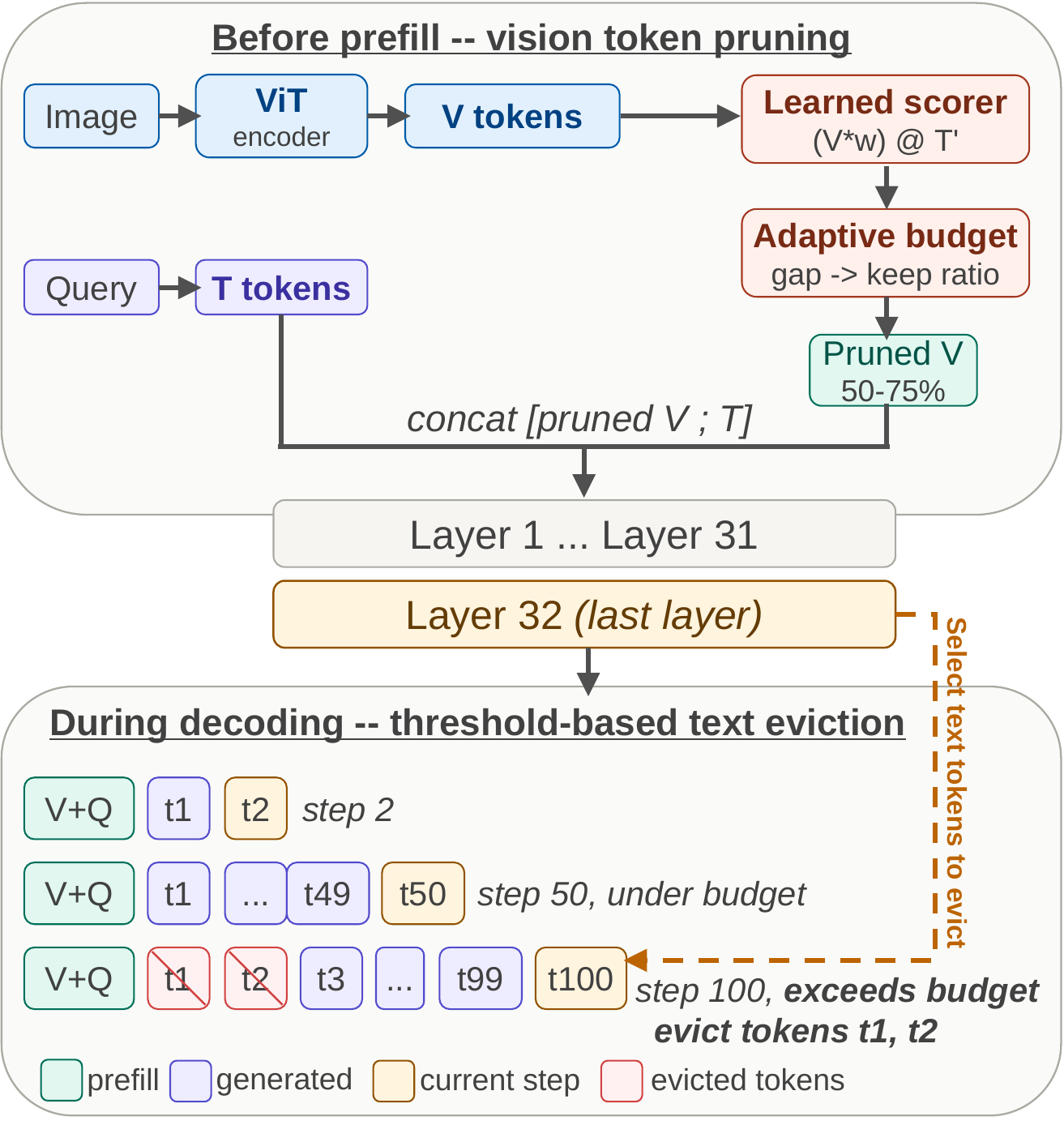}
  \end{center}
  \caption{Overview of AsymVLM. Vision token pruning operates in two-stage before prefill: (1) a learned scorer ranks token by cross-modal similarity, and (2)  per-sample adaptive budget determines the keep ratio based on the importance gap. During decoding, text tokens are evicted when they exceed a fixed budget, guided by last-layer attention.}
\end{wrapfigure}

These observations lead to a natural question: \textit{should vision and text tokens share the same compression strategy?} We argue that the answer is {\em no.} Vision token compression is fundamentally an \textit{input selection} problem, where we need to identify which parts of the image are needed to answer the query. In comparison, text token compression is a \textit{cache management} problem, where the model needs to decide which past tokens can be safely forgotten during generation, and is best addressed  during decoding.

In this paper, we present and evaluate {\bf AsymVLM}, an asymmetric token pruning framework for efficient VLM inference that decouples vision and text token compressions into modality-specific strategies. For vision tokens, we introduce a learned importance scorer to minimize the discrepancy between pruned and unpruned VLM outputs, yielding more faithful importance rankings than semantic similarity alone. We further propose a per-sample adaptive budget mechanism that dynamically adjusts the pruning ratio according to each input's importance distribution across vision tokens. This two-stage strategy is applied prior to the transformer layers, ensuring that all attention heads and layers operate on a consistent and query-relevant token set. For text tokens, we apply lightweight KV-cache eviction only when the generated sequence exceeds a predefined budget, thereby preserving the causal dependencies essential for coherent language generation.  Our {\bf contributions} in this work can be summarized as follows: 
 
\begin{itemize}[itemsep=5pt, parsep=0pt, topsep=2pt, partopsep=0pt, leftmargin=*]

\item We provide a systematic analysis revealing the fundamental asymmetry between vision and text tokens in structure, quantity, and compressibility for VLMs, and argue that modality-aware compression is essential for efficient VLM inference.

\item We propose AsymVLM, a dual-strategy framework that combines query-aware vision token pruning, before the transformer, with budget-constrained text token eviction during decoding.

\item We introduce a learned cross-modal importance scorer with per-sample adaptive budgeting, which \textit{dynamically} adjusts vision token retention based on each input's importance gap distribution, achieving consistent gains over uniform pruning at the same budget.

\item We conduct extensive experiments across multiple benchmarks, demonstrating that AsymVLM achieves the highest FLOPs savings (up to 54\%) among state-of-the-art methods while retaining competitive accuracy with competitive accuracy at 50\% vision token retention, and that our text eviction substantially outperforms standared LLM cache methods on free-form generation by adapting to the short-context nature of VLM inference.

\end{itemize}
\vspace{-10pt}

\section{Background and Related Work}
\vspace{-3pt}

\subsection{Vision-Language Model Inference} 
\vspace{-5pt}

Modern VLMs combine a pretrained vision encoder~\citep{zhai2023siglip,radford2021clip}, a modality projection layer, and an LLM backbone~\citep{liu2023llava, abdin2024phi3, bai2023qwenvl}. The vision encoder extracts patch features $\mathbf{Z} \in \mathbb{R}^{N \times D_v}$, which a projector maps to the LLM embedding space as visual tokens $\mathbf{V} \in \mathbb{R}^{N \times D}$. These are concatenated with text tokens $\mathbf{T} \in \mathbb{R}^{L \times D}$ to form the input $\mathbf{X} = [\mathbf{V}; \mathbf{T}]$. During \textit{prefill phase}, the model computes attention over all $N + L$ tokens and caches key-value pairs at each layer. During \textit{decoding phase}, tokens are generated autoregressively, each attending to the full KV cache. The prefill phase is ``compute-bound'' due to the quadratic attention cost over the input, while decoding is ``memory-bound'', as each step loads the entire KV cache from GPU memory.



\vspace{-5pt}
\subsection{Visual Token Reduction for VLMs}
\vspace{-5pt}
Existing methods prune vision tokens either within the LLM layers or before them. For example, FastV \citep{chen2024fastv} prunes low-attention tokens after layer 2; SparseVLM \citep{zhang2024sparsevlm} progressively prunes across layers using text-guided attention. Both access internal attention patterns, and are incompatible with FlashAttention \citep{dao2023flashattention}. VisPruner \citep{zhang2025vispruner} and FlashVLM \citep{cai2025flashvlm} prune between the vision encoder and LLM using visual saliency or cross-modal similarity, so the LLM only processes kept tokens. Despite their effectiveness, these approaches share two key limitations. First, they all apply a \textit{fixed pruning ratio} uniformly across inputs, ignoring the substantial variation in per-sample difficulty. Second, they score tokens based on \textit{semantic similarity} to the query or visual saliency, without directly accounting for how pruning affects the model's output. We address both limitations with an output-aware scorer and per-sample adaptive budgeting.

\vspace{-5pt}
\subsection{KV Cache Compression for LLMs}
\vspace{-5pt}
A complementary line of work {\em compresses} the KV cache during decoding in text-only LLMs by discarding low-importance entries. For instance, StreamingLLM \citep{xiao2024streamingllm} retains a fixed window of recent tokens plus attention sinks; H2O \citep{zhang2023h2o} evicts the tokens with the lowest accumulated attention; PyramidKV \citep{cai2024pyramidkv} allocates progressively smaller budgets to deeper layers; and DuoAttention \citep{xiao2024duoattention} enables per-head eviction by classifying heads into retrieval and streaming types. These approaches assume that all tokens are of the same modality and share similar properties in redundancy and temporal structure, and \citet{tu2024vlcache} confirms that they produce a suboptimal result when applied to VLMs {\em without} modality-aware adaptation.




\vspace{-5pt}
\section{Method}
\label{sec:method} 
\vspace{-5pt}

We first motivate our design choices through empirical analysis (Section~\ref{sec:analysis}), and then present AsymVLM, an asymmetric token pruning framework with three components: a learned importance scorer for vision tokens (Section~\ref{sec:scorer}), a per-sample adaptive budget (Section~\ref{sec:adaptive}), and a threshold-based text token eviction strategy (Section~\ref{sec:text_eviction}). Figure~\ref{fig:overview} shows the overview of AsymVLM. 

\vspace{-5pt}
\subsection{Asymmetric Analysis of Vision and Text Tokens}
\label{sec:analysis}
\vspace{-5pt}

We start by presenting three observations that reveal why vision and text tokens demand fundamentally different compression strategies.

\textbf{Observation 1: Vision and text tokens differ in structure, quantity, and compressibility.}  Vision tokens are produced by the vision encoder in a single forward pass, and all $N$ vision tokens are computed independently, without causal dependency among them. Removing a vision token does not invalidate any other vision token's representation. In contrast, text tokens are generated autoregressively, where each token $t_i$ is conditioned on all preceding tokens $t_1, \dots, t_{i-1}$ through the causal attention mask. This strict sequential dependency means that evicting a text token risks corrupting the representations of all subsequent tokens in the chain. We summarize the key differences between the two modalities in a detailed structural comparison in ~\ref{app:token_asymmetry}. 

In single-turn settings, vision tokens can account for over 95\% of the input (e.g., 2{,}500 vs.\ 30 text tokens in Phi-3.5-Vision), making them the dominant bottleneck and text compression unnecessary. However, free-form generation and multi-turn dialogue can produce hundreds of text tokens that accumulate in the KV cache, particularly on models with compact vision encoders (e.g., Gemma~3 with only 256 vision tokens per image). In these scenarios, the bottleneck shifts from vision during prefill to text during decoding, motivating our asymmetric design: \textit{aggressive vision pruning before the LLM  and threshold-based text eviction during decoding.}


\begin{figure*}[t]
\centering
\includegraphics[width=\textwidth]{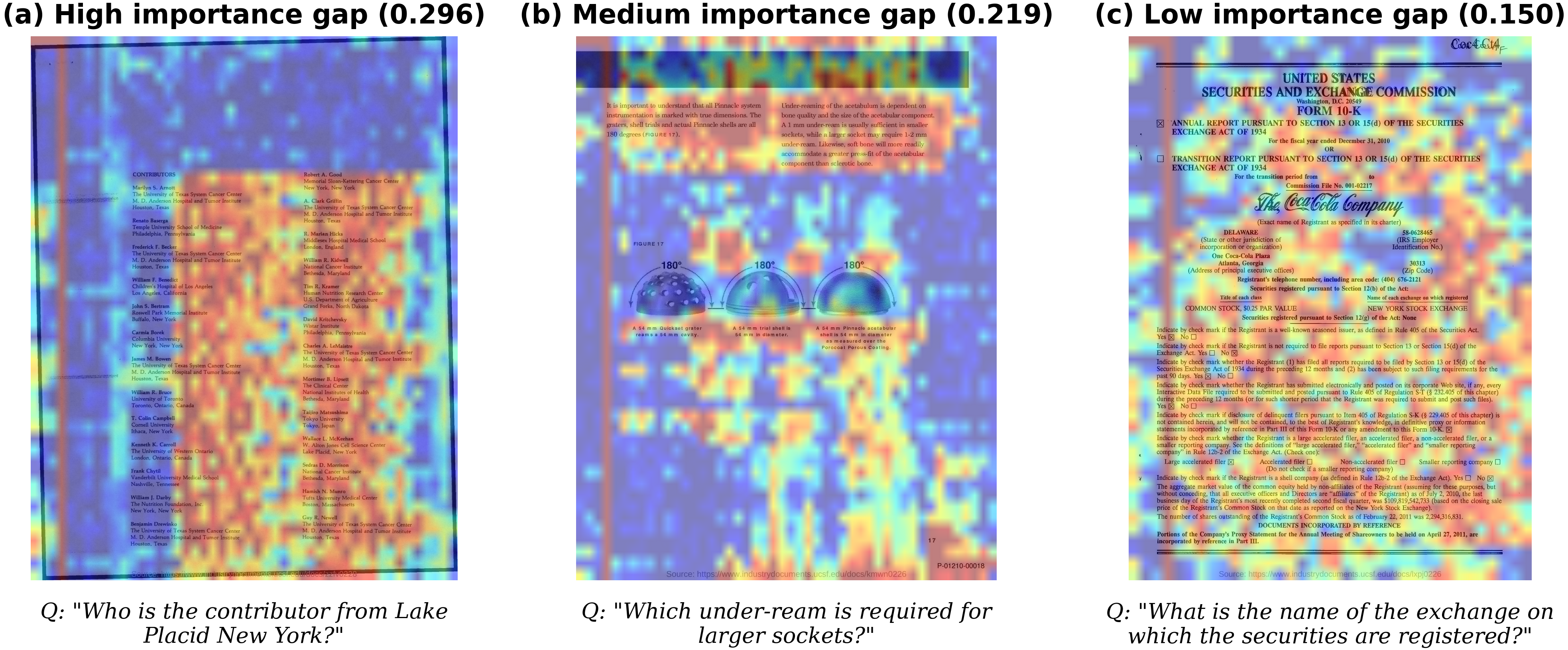}

\caption{Vision token importance heatmaps on three DocVQA samples with differing importance gaps. (a) High gap (0.296): localized importance, safe to prune aggressively. (b) Medium gap (0.219): split across text and diagrams. (c) Low gap (0.150): uniform importance, conservative retention needed.}

\label{fig:adaptive_examples_heatmap}
\vspace{-15pt}
\end{figure*}

\textbf{Observation 2: Existing vision token pruning relies solely
on input-side signals.} Current methods score tokens by visual saliency \citep{zhang2025vispruner}, cross-modal similarity \citep{cai2025flashvlm}, or attention patterns \citep{chen2024fastv, zhang2024sparsevlm}, without considering how removal affects the model's output. What ultimately matters is not how relevant a token \textit{appears}, but how much the output \textit{changes} when it is removed. This motivates our learned scorer (Section~\ref{sec:scorer}), which optimizes token
ranking with respect to output preservation.


\textbf{Observation 3: The importances of vision tokens vary substantially between samples.} Even with an effective vision token scorer, a fixed pruning ratio is suboptimal because task-relevant information is distributed differently across inputs. We define the importance gap as the difference between the 75th- and 25th-percentile importance scores; a large gap indicates clear separation between essential and dispensable tokens, while a small gap suggests uniform importance. Figure~\ref{fig:adaptive_examples_heatmap} illustrates three DocVQA examples: (a) ``Who is the contributor from Lake Placid, New York?'' concentrates importance on a single entry (gap = 0.296), allowing aggressive pruning; (b) ``Which under-ream is required for larger sockets?'' splits importance across text and diagrams (gap = 0.219), requiring moderate retention; and (c) ``What is the name of the exchange on which the securities are registered?'' spreads importance uniformly across a dense regulatory document (gap = 0.150), where aggressive pruning risks removing the relevant passage. This variation motivates our per-sample adaptive budget (Section~\ref{sec:adaptive}).

The gap distribution across 500 DocVQA samples approximately follows a normal distribution (mean 0.218, std 0.047; ~\ref{app:gap_distribution} shows the gap distribution figure), confirming that a fixed pruning ratio {\em cannot} accommodate this spectrum of input difficulty. This observation motivates our per-sample adaptive budget mechanism, described in Section~\ref{sec:adaptive}. 

\vspace{-5pt}
\subsection{Learned Vision Token Scorer}
\vspace{-5pt}
\label{sec:scorer}
As established in Observation 2, existing pruning methods rely on input-side signals that do not account for how token removal affects the model's output. We address this by training (using representative set of inputs) a lightweight {\em importance scorer} that learns to rank vision tokens based on their contributions to preserving the model's output representations, as shown in the top of Figure~\ref{fig:overview}.

Our scorer builds upon cross-modal similarity but introduces a learned per-dimension weighting that re-scales embedding dimensions to better align the scoring with output preservation. Given vision token embeddings $\mathbf{V} \in \mathbb{R}^{N \times D}$ and text token embeddings $\mathbf{T} \in \mathbb{R}^{L \times D}$ extracted from the model's embedding layer, we first $\ell_2$-normalize both to obtain $\hat{\mathbf{V}}$ and $\hat{\mathbf{T}}$. The similarity matrix and per-token importance scores are then computed as: 
\begin{equation}
\mathbf{S} = \left(\hat{\mathbf{V}} \odot w\right)
  \hat{\mathbf{T}}^\top \in \mathbb{R}^{N \times L}, \quad
  s_i = \max_{j} \; \mathbf{S}_{i,j},
  \label{eq:importance}
\end{equation}
where $w \in \mathbb{R}^{D}$ is a ``learnable'' weight vector and $\odot$ denotes element-wise multiplication (broadcast across tokens). The weight $w$ re-scales the normalized vision embeddings before computing the dot product with the normalized text embeddings, allowing the scorer to amplify dimensions that are informative for output preservation and suppress those that are not. The $\max$ aggregation selects each vision token's strongest alignment to any text token, capturing query relevance regardless of which specific token is most related. When $w = \mathbf{1}$, this reduces to the standard cosine similarity, making the raw cross-modal similarity a special case of our formulation.

The scorer is trained to minimize the discrepancy between the model's output under pruning and the unpruned baseline and we employ a soft score masking mechanism that allows gradients to flow through the pruning decision. Details of the masking mechanism is provided in ~\ref{app:scorer}. For each training sample, we perform two forward passes through the frozen VLM: a baseline pass with no masking, and a masked pass with the score mask applied. The training loss is the mean squared error between the last hidden states at text token positions:
\begin{equation}
  \mathcal{L} = \frac{1}{L} \sum_{i=1}^{L}
  \left\| h_i^{\text{masked}} - h_i^{\text{baseline}} \right\|_2^2
  + \lambda \left\| w - \mathbf{1} \right\|_2^2
  \label{eq:loss}
\end{equation}
where $h_i^{\text{baseline}}$ and $h_i^{\text{masked}}$ are the last-layer hidden states at text position $i$ from the baseline and masked forward passes, respectively, and the second term is an $\ell_2$ regularizer ($\lambda = 0.001$) that keeps $w$ close to $\mathbf{1}$, preventing the scorer from
diverging far from the cosine similarity baseline. Only $w$ is updated during training, and all model parameters remain frozen. The loss is computed exclusively over text token positions, as
these directly determine the generated output.


\vspace{-5pt}
\subsection{Per-Sample Adaptive Budget}
\label{sec:adaptive}
\vspace{-5pt}
As shown in Observation 3, the importance gap varies substantially across samples, making a fixed pruning ratio inherently suboptimal. We propose a per-sample adaptive budget mechanism that adjusts the vision token keep ratio based on each input's importance gap. The intuition is straightforward: when importance scores are uniformly distributed across vision tokens, no clear subset can be safely discarded, and the model should retain more tokens. When scores are highly distinguishable with a clear separation between important and unimportant tokens, aggressive pruning is safe. As shown in Figure~\ref{fig:overview}, the adaptive budget is applied after learned scorer.

For each input, we compute the importance gap $g$ from the scorer outputs $\{s_i\}_{i=1}^{N}$ as the difference between the $k$-th highest and $k$-th lowest importance scores:   
\(\displaystyle g=s^{(k)}-s^{(N-k+1)}\), 
where $k = \lfloor N/4 \rfloor$ and $s^{(k)}$ denotes the $k$-th largest value in the score sequence. This measures the spread between the boundaries of the top and bottom quartiles. A large $g$ indicates a clear distinction between essential and dispensable tokens, while a small $g$ indicates a uniform distribution in which pruning carries greater risk. Figure~\ref{fig:gap_distribution} shows the gap distribution across 500 DocVQA samples, which approximately follows a normal distribution (mean 0.218, std 0.047). 

Given the importance gap $g$, we map it to a per-sample keep ratio $r(g)$ that determines the fraction of vision tokens to retain. A smaller gap should yield a higher keep ratio (more conservative), while a larger gap permits a lower keep ratio (more aggressive). We explore two approaches:

\textbf{Method A: Threshold sweep.}
We partition samples into three difficulty tiers using two gap thresholds $g_{\text{lo}}$ and $g_{\text{hi}}$, each assigned a
fixed keep ratio:
\begin{equation}
  r(g) =
  \begin{cases}
    r_{\text{conservative}}, & \text{if } g < g_{\text{lo}} \\
    r_{\text{moderate}},     & \text{if } g_{\text{lo}} \leq g < g_{\text{hi}} \\
    r_{\text{aggressive}},   & \text{if } g \geq g_{\text{hi}}
  \end{cases}
  \label{eq:threshold_sweep}
\end{equation}
where $r_{\text{conservative}} > r_{\text{moderate}} > r_{\text{aggressive}}$. The thresholds and ratios are selected via grid search over a small calibration subset.

\textbf{Method B: Learned linear mapping.}
Instead of discrete tiers, we learn a continuous mapping from gap to keep ratio:
\(\displaystyle
r(g) = \mathrm{clamp}\!\left(
r_{\text{target}} + a(g - \bar{g}),\;
r_{\min},\;
r_{\max}
\right)
\).


where $r_{\text{target}}$ is the desired average keep ratio,  $\bar{g}$ is the mean gap computed from the calibration set, and $a$ is a learned slope optimized on the calibration subset to maximize task accuracy. A negative $a$ ensures that samples with larger gaps receive lower keep ratios and samples with smaller gaps receive higher keep ratios. The output is clamped to prevent extreme pruning ratios.

Both methods yield similar improvements over uniform pruning, with consistent gains across different average budget levels. We evaluate both approaches in Section~\ref{sec:experiments} and adopt Method A as the default due to its simplicity and transparency.

\vspace{-5pt}
\subsection{Threshold-Based Text Token Eviction}
\label{sec:text_eviction}
\vspace{-5pt}

While vision token pruning addresses the prefill bottleneck, text tokens can become the dominant memory consumer during free-form generation and multi-turn dialogue, particularly on models with compact vision encoders. We apply a threshold-based text eviction strategy: \textit{text tokens are left untouched until they exceed a fixed budget}, at which point the lowest-importance generated tokens are evicted. When eviction is triggered, importance is estimated differently depending on the available context. In single-turn generation, we score generated tokens by their \textit{attention received at the final transformer layer} of the current decoding step, averaged across heads.  In multi-turn dialogue, where the memory pressure comes from accumulated past responses rather than a single long generation, we score entire previous answers by cross-modal similarity to the current images and evict at the turn-level, preserving all questions and images.

As shown in the bottom of Figure~\ref{fig:overview}, our design is tailored to the short-context nature of VLM inference. Unlike long-context LLMs, VLM text generation is typically brief (under a few hundred tokens), questions across turns are largely independent, and reasoning is primarily grounded in visual content rather than long-range textual dependencies. These properties mean that (1) the current step's final-layer attention is a sufficient importance signal without cumulative multi-layer tracking, and (2) prefill tokens can be unconditionally protected since generated tokens remain few enough to serve as eviction candidates. Standard methods such as H2O \citep{zhang2023h2o} and StreamingLLM \citep{xiao2024streamingllm} are designed for long-context scenarios where cumulative multi-step, multi-layer attention tracking is necessary because a token's relevance may shift substantially over the course of generation. They also treat all tokens uniformly, which risks evicting vision or prompt tokens that carry irreplaceable visual context.


\vspace{-10pt}
\section{Experiments}
\label{sec:experiments}
\vspace{-5pt}

\subsection{Experimental Setup}
\vspace{-5pt}
\textbf{Models.}
We evaluate AsymVLM on two VLM architectures with contrasting vision encoder designs: (1) Phi-3.5-Vision 4B \citep{abdin2024phi3}, which produces up to 2{,}500 vision tokens per image using a multi-crop strategy, making vision tokens the dominant memory consumer; and (2) Gemma~3 12B \citep{team2025gemma3}, which uses a compact vision encoder generating only 256 tokens per image, shifting the bottleneck toward text tokens in long-generation scenarios. This pairing allows us to evaluate AsymVLM under both vision-dominant and text-dominant regimes. The learned importance scorer is trained on 1{,}500 samples from the DocVQA (training details in ~\ref{app:training}) .

\begin{table}[t]
\centering
\caption{Token composition across evaluation scenarios. The dominant modality shifts based on vision encoder compactness, response length, and conversation depth.}
\label{tab:token_composition}
\small
\begin{tabular}{lrrrcc}
\toprule
\textbf{Scenario} & \textbf{Avg Vision} & \textbf{Avg Query} & \textbf{Generated} & \textbf{Vis.\%} & \textbf{Txt.\%} \\
\midrule
\multicolumn{6}{l}{\textit{Phi-3.5-Vision 4B}} \\
MME &  737 & 30 & 3--5    & 96.1\% & 3.9\%  \\
 DocVQA & 2231 & 36  &   5--15    & 98.4\% & 1.6\%  \\
 OCRVQA &  2312 & 30  &  3--8  & 98.7\% & 1.3\%  \\
 TextVQA & 2044 &  31 &  3--10  & 98.5\% & 1.5\%  \\
 ChartQA &  2236  & 37  &  5--15  & 98.4\% & 1.6\%  \\
\midrule
\multicolumn{6}{l}{\textit{Gemma 3 12B}} \\
LLaVA-Bench
  & $\sim$256 & 100--200 & 350--400 & 37\% & 63\% \\
MMDU (last turn)
  & $\sim$1{,}280 & $\sim$500 & $\sim$2{,}900 & 27\% & 73\% \\
\bottomrule
\end{tabular}
\vspace{-10pt}
\end{table}

\textbf{Benchmarks.}
We evaluate on seven benchmarks: DocVQA \citep{mathew2021docvqa} (document QA, ANLS), ChartQA \citep{masry2022chartqa} (chart reasoning, relaxed accuracy) \footnote{For consistency, we evaluate on the full DocVQA dataset including the 1{,}500 samples used to train the scorer. Excluding these samples yields a near-identical score.}, TextVQA \citep{singh2019textvqa} (scene text QA, accuracy), MME \citep{fu2023mme} (Yes/No visual QA, accuracy), OCRVQA \citep{mishra2019ocrvqa} (book cover QA, accuracy), LLaVA-Bench \citep{liu2023llava} (free-form generation, GPT-4), and MMDU \citep{liu2024mmdu} (multi-turn multi-image dialogue, GPT-4o). The first five produce concise outputs and are evaluated on Phi-3.5-Vision; LLaVA-Bench and MMDU involve verbose generation and are evaluated on Gemma~3.


\textbf{Strategy selection.}
A key design principle of AsymVLM is that the compression strategy adapts to where the bottleneck is, which shifts according to the vision encoder, the response length, and the conversation depth. Table~\ref{tab:token_composition} shows the token composition across our evaluation scenarios. For short-answer QA (MME, DocVQA, OCRVQA) on Phi-3.5-Vision, vision tokens account for approximately 97\% of total tokens, making vision pruning the primary strategy. For free-form generation (LLaVA-Bench and MMDU), text dominates on Gemma~3, calling for text eviction instead. AsymVLM applies the appropriate strategy per scenario rather than uniform compression across both modalities.

\textbf{Baselines.}
For vision token pruning, we compare with FastV \citep{chen2024fastv}, which prunes low-attention vision tokens after layer 2, and SparseVLM \citep{zhang2024sparsevlm}, which progressively prunes across layers using text-guided attention.  For text token eviction, we compare with H2O \citep{zhang2023h2o}, which evicts tokens with the lowest accumulated attention at every decoding step, and StreamingLLM \citep{xiao2024streamingllm}, which retains only a fixed window of recent tokens plus a small set of initial attention sink tokens. All experiments are run on a single NVIDIA A100 GPU with FP16 inference

\vspace{-5pt}
\subsection{Main Results}
\vspace{-5pt}
We first compare AsymVLM against existing vision token pruning methods on five benchmarks using Phi-3.5-Vision. Table~\ref{tab:main_results} reports results with three keep ratios (75\%, 65\%, 50\%). It can be observed that AsymVLM achieves the strongest results on document and chart understanding benchmarks, where visual information is spatially localized and query-specific. On DocVQA, AsymVLM outperforms FastV by 2.2\% at 75\% retention (85.75 vs.\ 83.53) and by 3.3\% at 65\% (85.07 vs.\ 81.74). On ChartQA, AsymVLM maintains or exceeds the full-token baseline even under compression (68.0 and 67.8 at 75\% and 65\% vs.\ 67.4 baseline). Compared to SparseVLM, AsymVLM achieves higher accuracy on DocVQA and ChartQA at all keep ratios while performing comparably on the remaining benchmarks. These improvements are attributable to the output-aware scorer, which {\em learns} to preserve the specific vision tokens that influence the model's answer rather than relying on attention patterns that may not reflect true importance for text-extraction tasks. On holistic understanding tasks (MME, TextVQA, OCRVQA), where token importance is more uniformly distributed, all methods perform similarly, with AsymVLM maintaining a slight edge on MME and TextVQA.

\begin{table}[t]
\centering
\caption{Main results on Phi-3.5-Vision across five benchmarks at various vision token keep ratios. We report ANLS (\%) for DocVQA, relaxed accuracy (\%) for ChartQA, accuracy (\%) for MME, TextVQA, and OCRVQA. Best results at each keep ratio are in \textbf{bold}.}
\label{tab:main_results}
\small
\begin{tabular}{llcccccc}
\toprule
\textbf{Keep ratio} & \textbf{Method}
& \textbf{DocVQA} & \textbf{ChartQA} & \textbf{MME}
& \textbf{TextVQA} & \textbf{OCRVQA}
& \textbf{FLOPs Saved} \\
\midrule
100\% & Baseline
  & 85.91 & 67.40 & 80.90 & 73.54 & 73.0 & --- \\
\midrule
\multirow{3}{*}{75\%}
  & AsymVLM  & \textbf{85.75} & \textbf{68} & \textbf{81.60} & \textbf{73.78} & 72.40 & \textbf{28\%} \\
  & FastV$^\dagger$ & 83.53 & 66 & 81.40 & 73.48 & 72.60 & 0\% \\
  & SparseVLM$^\ddagger$ & 84.58 & 67.40 & 81.40 & 73.66 & \textbf{72.80} & 18\% \\
\midrule
\multirow{3}{*}{65\%}
  & AsymVLM  & \textbf{85.07} & \textbf{67.80} & \textbf{81.50} & \textbf{73.45} & 72.0 & \textbf{39\%} \\
  & FastV$^\dagger$ & 81.74 & 66 & 81.20 & 72.66 & 72.60 & 0\% \\
  & SparseVLM$^\ddagger$ & 84.40 & 67 & 81.40 & 73.34 & \textbf{72.60} & 35\% \\
\midrule
\multirow{3}{*}{50\%}
  & AsymVLM  & 83.22 & \textbf{66.40} & \textbf{81.30} & \textbf{73.20} & 71.60 & \textbf{54\%} \\
  & FastV$^\dagger$ & 77.30 & 61.40 & 81.20 & 72.24 & 72.60 & 0\% \\
  & SparseVLM$^\ddagger$ & \textbf{83.38} & 65.80 & 81.20 & \textbf{73.20} & \textbf{72.60} & 47\% \\
\bottomrule
\multicolumn{8}{l}{\footnotesize $^\dagger$ FastV masks attention
without reducing sequence length; no memory or KV cache savings.} \\
\multicolumn{8}{l}{\footnotesize $^\ddagger$ SparseVLM prunes progressively but early layers process all tokens; no peak memory or KV cache savings.} \\
\multicolumn{8}{l}{\footnotesize AsymVLM prunes before the transformer, achieving 44\%--75\% peak memory reduction and 25\%--50\% KV cache reduction.}
\end{tabular}
\vspace{-20pt}
\end{table}

Since AsymVLM physically removes vision tokens before the transformer, it achieves the highest FLOPs \footnote{ $4nd^2 + 2n^2d + 3ndm$ (QKV projection + attention + FFN), where $n$ is the sequence length, $d$ the hidden dimension, and $m$ the FFN dimension} savings among all compared methods: 28\%, 39\%, and 54\% at keep ratios of 75\%, 65\%, and 50\% respectively. FastV applies attention masking without reducing the sequence length, yielding  no FLOPs or memory savings. SparseVLM achieves moderate savings in FLOPs through progressive pruning, but remains below AsymVLM at every keep ratio (e.g., 47\% vs.\ 54\% at 50\% retention). Moreover, both FastV and SparseVLM maintain the same peak memory and KV cache size as the unpruned baseline, since their early layers still process all tokens. AsymVLM is the only method that reduces peak memory (to as low as 25\% at 50\% retention) and KV cache size proportionally, enabling inference on memory-constrained GPUs that cannot run the full model. 

Notably, the learned importance scorer is trained on only 1{,}500 samples from the DocVQA training set, but it generalizes effectively across all five benchmarks without any task-specific tuning. On ChartQA and TextVQA, which involve different visual layouts and question types from the training data, AsymVLM still achieves consistent improvements. This cross-task generalization suggests that the per-dimension weighting captures general principles about which embedding dimensions are informative for output preservation, rather than overfitting to document-specific patterns. 

When text tokens dominate the sequence, as in free-form generation on Gemma~3, vision pruning offers limited savings and text eviction becomes the relevant strategy. We compare our threshold-based  text eviction against H2O~\citep{zhang2023h2o} and StreamingLLM~\citep{xiao2024streamingllm} on both LLaVA-Bench (single-turn) and MMDU (multi-turn) with the eviction threshold set to 90 tokens for LLaVA-Bench and 512 tokens for MMDU. As shown in Table~\ref{tab:text_eviction},  our method maintains stable performance across all retention levels (75.69 / 75.33 / 74.69) on LLaVA-Bench, while H2O degrades more noticeably (75.3 $\to$ 69.83) and StreamingLLM drops
severely (58.33 $\to$ 53.0). The robustness of our approach stems from protecting all prefill tokens and relying on the current step's attention at the final layer, which provides a sufficient importance signal for VLMs' short generated sequences without the overhead of cumulative multi-layer tracking. On MMDU, our approach substantially outperforms both baselines, achieving 7.03 at 90\% retention compared to 5.61 for both H2O and StreamingLLM. Eviction at 50\% retention matches the baseline exactly (6.80), and eviction at 90\% even slightly improves the score (+0.23), suggesting that older generated responses in multi-turn dialogue introduce noise that eviction beneficially removes. H2O and StreamingLLM degrade dramatically (to 5.02 and 4.95 at 50\%) because they treat all tokens as eviction candidates, including vision and prompt tokens that carry irreplaceable context for subsequent turns. 

\vspace{-5pt}
\subsection{Ablation Studies}
\vspace{-5pt}
\textbf{Learned Scorer vs. Cosine Similarity}
We compare our learned importance scorer against two baselines: spiral ranking (position-based, evicting edge tokens first) and raw cosine similarity between vision and text embeddings. All methods are evaluated with various vision token retention on Phi-3.5-Vision.  As shown in Table~\ref{tab:scorer_ablation}, the learned scorer substantially outperforms both baselines on text-detection tasks. On DocVQA, it achieves 83.22\% compared to 68.04\% for cosine similarity and 55.39\% for spiral ranking with 50\% of vision tokens , a 16-point improvement over cosine similarity by learning which embedding dimensions are most informative for output preservation. In OCRVQA, the improvement is similar (73.20\% vs.\ 56.60\%). On MME, all scoring methods perform equally well, as this benchmark consists of coarse Yes/No questions where even naive pruning preserves sufficient visual information. 

\begin{table}[t]
\centering
\caption{Text eviction comparison on Gemma~3. LLaVA-Bench: GPT-4 score (/100). MMDU: GPT-4o score (/10).}
\label{tab:text_eviction}.
\small
\begin{tabular}{lcccccccccc}
\toprule
& & \multicolumn{3}{c}{\textbf{H2O}} & \multicolumn{3}{c}{\textbf{StreamingLLM}} & \multicolumn{3}{c}{\textbf{AsymVLM}} \\
\cmidrule(lr){3-5} \cmidrule(lr){6-8} \cmidrule(lr){9-11} \textbf{Dataset} & \textbf{Baseline}
& 90\% & 75\% & 50\%
& 90\% & 75\% & 50\%
& 90\% & 75\% & 50\% \\
\midrule
LLaVA-Bench & 79 & 75.3 & 74.5 & 69.83 & 58.33 & 56 & 53 & \textbf{75.69} & \textbf{75.33} & \textbf{74.69} \\
MMDU & 6.80 & 5.61 & 5.46 & 5.02 & 5.61 & 5.24 & 4.95 & \textbf{7.03} & \textbf{7.00} & \textbf{6.80} \\
\bottomrule
\end{tabular}
\vspace{-5pt}
\end{table}

\textbf{Adaptive vs. Uniform Budget}
The results given in Table~\ref{tab:main_results} indicate that the learned scorer achieves near-lossless performance at 75\% vision token retention on DocVQA ($-$0.38\% from baseline). From this ``sweet spot'', we investigate whether per-sample adaptive budgeting can push the pruning ratio further below 75\% while minimizing accuracy loss. We compare adaptive budgeting against uniform pruning with the learned scorer at various average keep ratios on DocVQA. 

\begin{table}[t]
\centering
\caption{Comparison of vision token scoring methods at various keep ratios on Phi-3.5-Vision. We report ANLS (\%) for DocVQA and accuracy (\%) for OCRVQA and MME.}
\label{tab:scorer_ablation}
\small
\begin{tabular}{llcccc}
\toprule
\textbf{Dataset} & \textbf{Keep ratio}
& \textbf{Spiral} & \textbf{Cross-sim}
& \textbf{Learned} & \textbf{Full tokens} \\
\midrule
\multirow{3}{*}{DocVQA}
  & 75\% & 75.81& 79.48 & \textbf{85.75} & \multirow{3}{*}{85.91} \\
  & 65\% & 69.38 & 71.65 & \textbf{85.07} & \\
  & 50\% & 55.39 & 68.04 & \textbf{83.22}& \\
\midrule
\multirow{3}{*}{OCRVQA}
  & 75\% & 64.8 & 69.20 & \textbf{73.78} & \multirow{3}{*}{73.00} \\
  & 65\% & 60 &  65.40& \textbf{73.45} & \\
  & 50\% & 51.80 & 56.60 & \textbf{73.20} & \\
\midrule
\multirow{3}{*}{MME}
  & 75\% & 80 & 80.40& \textbf{81.60} & \multirow{3}{*}{80.90} \\
  & 65\% & 80 & 80.80  & \textbf{81.50} & \\
  & 50\% & 79.60 & 80.90 & \textbf{81.30} & \\
\bottomrule
\end{tabular}
\vspace{-15pt}
\end{table}

Table~\ref{tab:adaptive_ablation} shows that both adaptive methods consistently outperform uniform vision pruning based on scorer at all budget levels. The threshold sweep (Method A) achieves gains of 1.8--3.8\% by allocating more tokens to low-gap samples (where importance is uniformly distributed) and fewer tokens to high-gap samples (where a clear focal point exists). The learned linear mapping (Method B) yields comparable gains with a continuous mapping. At an average budget of 65\%, the threshold sweep achieves 84.35\% ANLS, only 1.56\% below the full-token baseline (85.91\%).                                 

\begin{table}[H]
\vspace{-10pt}
\centering
\caption{Two adaptive budget methods on  DocVQA. Both methods consistently outperform uniform pruning at the same average budget.}
\label{tab:adaptive_ablation}
\small
\begin{tabular}{lccc}
\toprule
\textbf{Avg.\ budget} & \textbf{Uniform} & \textbf{Threshold sweep}
& \textbf{Learned linear mapping} \\
\midrule
70\% & 85.28 & \textbf{85.77} & 85.20  \\
65\% & 84.25 & \textbf{85.31}  & 84.35  \\
60\% & 83.22 & \textbf{83.83} &   83.60\\
55\% & 80.86 & 82.40  & \textbf{82.82} \\
50\% & 78.50& 80.30  & \textbf{81.20} \\
\bottomrule
\end{tabular}
\vspace{-10pt}
\end{table}

\vspace{-10pt}
\section{Conclusion and Limitations}
\vspace{-5pt}
We presented {\bf AsymVLM}, an asymmetric token pruning framework that treats vision and text tokens as fundamentally different compression targets. By combining a learned output-aware scorer with per-sample adaptive budgeting for vision tokens and threshold-based eviction for text tokens, AsymVLM achieves the highest savings in FLOPs (up to 54\%) among compared methods, while outperforming existing approaches on  document and chart understanding tasks and maintaining competitive accuracy across all benchmarks. Our evaluation focused on Phi-3.5-Vision and Gemma~3; generalization to other architectures (e.g., LLaMA-based VLMs, Qwen-VL) remains to be validated. Furthermore, our text eviction strategy uses a fixed threshold that does not adapt to conversation difficulty, and its effectiveness on very long multi-turn dialogues (50+ turns) has not been tested.

\bibliographystyle{plainnat}
\bibliography{reference}


\appendix

\renewcommand{\thesection}{Appendix \Alph{section}}

\section{Token Asymmetry Analysis}
\label{app:token_asymmetry}

Table~\ref{tab:token_asymmetry} summarizes the structural differences between vision and text tokens in VLMs that motivate our asymmetric compression design.

\begin{table}[H]
\centering
\caption{Structural comparison of vision and text tokens in VLMs.}
\label{tab:token_asymmetry}
\small
\begin{tabular}{lll}
\toprule
\textbf{Property} & \textbf{Vision tokens} & \textbf{Text tokens} \\
\midrule
Origin & Vision encoder (parallel) & Tokenizer + autoregressive decoding \\
Dependency & None (spatially independent) & Strict causal order \\
Count & 1{,}500--2{,}500 per image & 20--250 (prompt + output) \\
Redundancy & High (spatial neighbors) & Low (each token matters) \\
Importance & Query-dependent, variable & Uniformly high \\
Bottleneck & Prefill (dominates input) & Decoding (accumulates over generation) \\
Strategy & Aggressive pre-transformer pruning & Progressive decode-time eviction \\
\bottomrule
\end{tabular}
\end{table}

\section{Importance Gap Distribution}
\label{app:gap_distribution}
\begin{figure}
  \centering
  
  \includegraphics[width=0.7\textwidth]{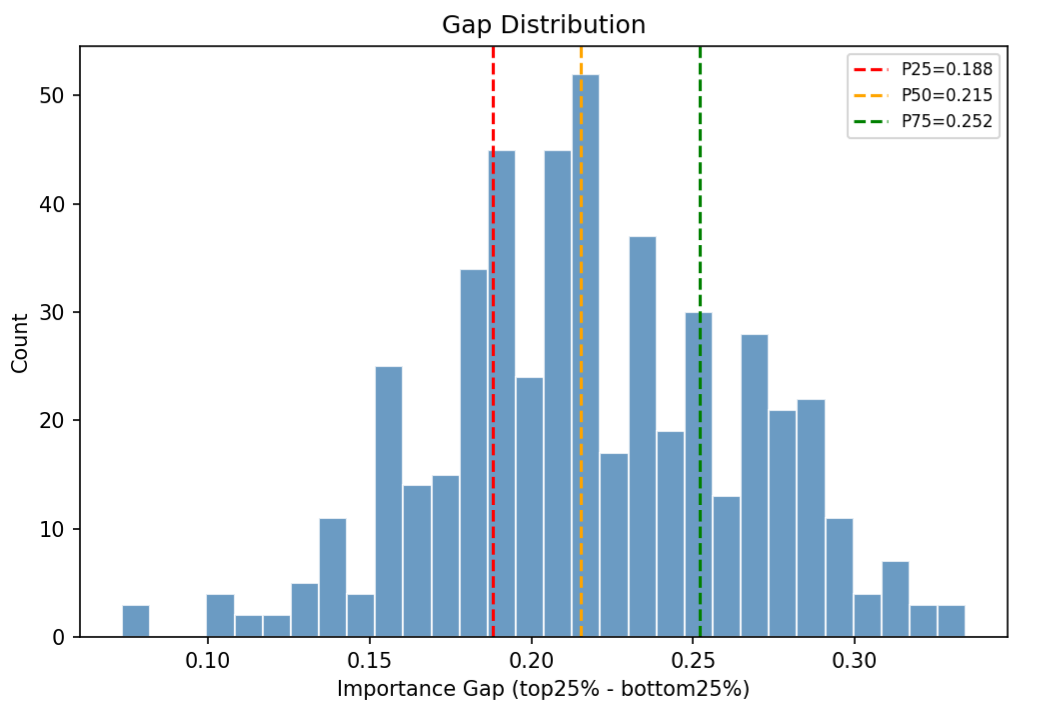}
  \caption{Importance gap distribution across 500 DocVQA samples.}
  \label{fig:gap_distribution}
  
\end{figure}

Figure~\ref{fig:gap_distribution} shows the importance gap distribution across 500 DocVQA samples (mean 0.218, std 0.047). The distribution is approximately normal, with the majority of samples falling between 0.15 and 0.28. Samples in the left tail (gap $<$ 0.15) have uniformly distributed token importance and require conservative pruning, while samples in the right tail (gap $>$ 0.28) have clear focal points and tolerate aggressive pruning.

\section{Score Masking}
\label{app:scorer}
Given the importance scores $\{s_i\}_{i=1}^{N}$ and a target keep ratio $r \in (0, 1]$, we compute a threshold $\theta$ as the importance score of the $\lfloor Nr \rfloor$-th highest scoring token. A continuous penalty is then added to the attention logits for each vision token at every layer:
\begin{equation}
  m_i = -C \cdot \text{sigmoid}\!\left(\frac{\theta - s_i}{\tau}\right)
  \label{eq:score_mask}
\end{equation}
where $C$ controls the masking strength and $\tau$ the sharpness of the transition. For tokens with $s_i \gg \theta$, the sigmoid outputs near zero and the penalty vanishes; for tokens with $s_i \ll \theta$, the sigmoid saturates near one and the token receives a penalty of approximately $-C$ in the attention logits, effectively suppressing attention to it after the softmax. The threshold $\theta$ is detached from the computation graph to prevent gradient flow through the top-$k$ operation, while gradients still flow through $s_i$ back to the scorer weights $w$. The mask $\mathbf{m} \in \mathbb{R}^{N+L}$  is constructed by assigning the penalty to vision-token positions and zero to text-token positions, and then incorporating it into the causal attention mask. Importantly, this mask is shared across all layers and attention heads, ensuring that every layer operates on a consistent representation of the visual input. Text token positions receive zero penalty.

\section{Learned Importance Scorer Training Details}
\label{app:training}

The learned importance scorer is trained on 1{,}500 samples from the DocVQA training set for 3 epochs using AdamW with a learning
rate of $10^{-3}$. The score masking uses $C = 5.0$ and $\tau = 1.0$. The learned weight $w$ is initialized to $\mathbf{1}$ and clamped to $[0, 3]$ after each update to prevent instability. The $\ell_2$ regularization weight is $\lambda = 0.001$. Training requires only a A100 GPU and completes within one hour, as each step involves two forward passes through the frozen model with no backward pass through the model parameters.

For the per-sample adaptive budget, the gap thresholds and keep ratios are selected via grid search over a 500-sample calibration
set from DocVQA validation. The learned linear mapping slope $a$ is optimized on the same calibration set.


\end{document}